\documentclass[12pt]{article}

\usepackage{arxiv}
\usepackage{graphicx}
\usepackage{subcaption}
\usepackage{color}
\usepackage{amsmath}
\usepackage{url}
\usepackage{hyperref}
\usepackage{amssymb}

\title{Fairness Index Measures to Evaluate Bias in Biometric Recognition}

\author{
Ketan~Kotwal\hspace{1mm}\mbox{\href{https://orcid.org/0000-0003-3766-0881}{\includegraphics[width=4mm]{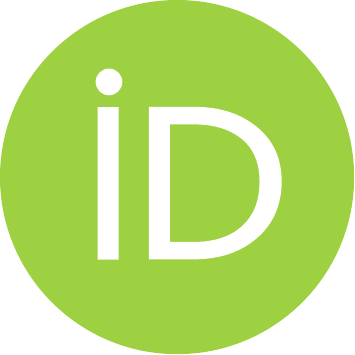}}}\\
Idiap Research Institute, Switzerland\\
\texttt{ketan.kotwal@idiap.ch}\\
\And
S\'{e}bastien Marcel\hspace{1mm}\mbox{\href{https://orcid.org/0000-0002-2497-9140}{\includegraphics[width=4mm]{images/orcid.pdf}}}\\
Idiap Research Institute, Switzerland\\
University of Lausanne, Switzerland\\
\texttt{ sebastien.marcel@idiap.ch}\\
}
\date{}

\hypersetup{
pdftitle={Fairness Index Measures},
pdfsubject={},
pdfauthor={Ketan Kotwal},
pdfkeywords={Biometrics, Demographic Fairness, Fairness Evaluation.},
}

\begin{document}
\maketitle

\begin{abstract}

The demographic disparity of biometric systems has led to serious concerns
regarding their societal impact as well as applicability of such systems in
private and public domains. A quantitative evaluation of demographic fairness
is an important step towards understanding, assessment, and mitigation of
demographic bias in biometric applications. While few, existing fairness
measures are based on post-decision data (such as verification accuracy) of
biometric systems, we discuss how pre-decision data (score distributions)
provide useful insights towards demographic fairness. In this paper, we
introduce multiple measures, based on the statistical characteristics of score
distributions, for the evaluation of demographic fairness of a generic
biometric verification system. We also propose dif{}ferent variants for each
fairness measure depending on how the contribution from constituent demographic
groups needs to be combined towards the final measure. In each case, the
behavior of the measure has been illustrated numerically and graphically on
synthetic data. The demographic imbalance in benchmarking datasets is often
overlooked during fairness assessment. We provide a novel weighing strategy to
reduce the effect of such imbalance through a non-linear function of sample
sizes of demographic groups. The proposed measures are independent of the
biometric modality, and thus, applicable across commonly used
biometric modalities (\textit{e.g.}, face, fingerprint, etc.).

\keywords{Biometrics \and Demographic \and Fairness \and Fairness Evaluation.}



\end{abstract}


\section{Introduction}
\label{sec:intro}
%

In recent years, algorithmic bias and fairness have emerged as noteworthy
challenges for automated biometric
systems~\cite{sun_pami,howard2019,damer_2020,nist3,raji2019actionable}. A
biometric system or algorithm is considered to be biased if significant
differences in its operation can be observed for different demographic groups of
individuals~\cite{damer_2020}. With growing adoption of various biometric
applications, the non-equitable performance of such applications across
demographic groups has led to several discussions and
debates~\cite{mit_review,harvard_review,un_review}. Several institutions have
conducted evaluations (w.r.t demographic bias) of popular biometric
applications, such as face recognition, developed by commercial
vendors~\cite{nist3,raji2019actionable,raji2020saving,cook_2019}. On academic
front as well, the research in understanding, estimating, and mitigating
demographic bias is gaining significant
traction~\cite{sun_pami,damer_2020,eccv2020fairface,garcia2019harms}.

As the issue of fairness in biometric systems has received attention lately,
very few attempts have been made to define fairness measures for generic and/or
specific biometric applications. As per Howard
\textit{et~al.}~\cite{howard2019}, the fairness of biometric system can be
measured using two approaches: differential performance and differential
outcome. The former approach refers to the difference in the genuine or imposter
distributions between specific demographic groups for a given biometric task,
whereas the latter deals with differences in classification error rates among
demographic groups. The differential performance is, thus, independent of any
classification threshold, while the differential outcomes are functions of a
chosen threshold that binarizes scores into match or no-match. 

A recently proposed fairness measure- Fairness Discrepancy Rate (FDR)- is based
on the differential outcome of biometric verification systems~\cite{fdr2021}. To
compute the FDR, authors first assess the maximum discrepancy in the false match
rate (FMR) and false non-match rate (FNMR) of different demographic groups for
several score thresholds. The fairness of the system is evaluated through a
weighted combination of these maximum discrepancies. The FDR has also been
adapted to measure the fairness in detection of face morphing attacks
in~\cite{fdr_morph}. Gong~\textit{et al.} have considered the area under the ROC
(Receiver Operating Characteristic) curve as a proxy to measure demographic
differentials~\cite{gong_debias}. In its special report on demographic effects
in face recognition~\cite{nist3}, the Face Recognition Vendor Test (FRVT) has
employed differential outcome-based strategy where they discuss the impact FMR
and FNMR using a global threshold. A demographic-specific score thresholding has
been analyzed in~\cite{cook_2019}.  

\begin{figure}[!b]
\centering
\begin{subfigure}[b]{0.4\textwidth}
    \includegraphics[width=\textwidth]{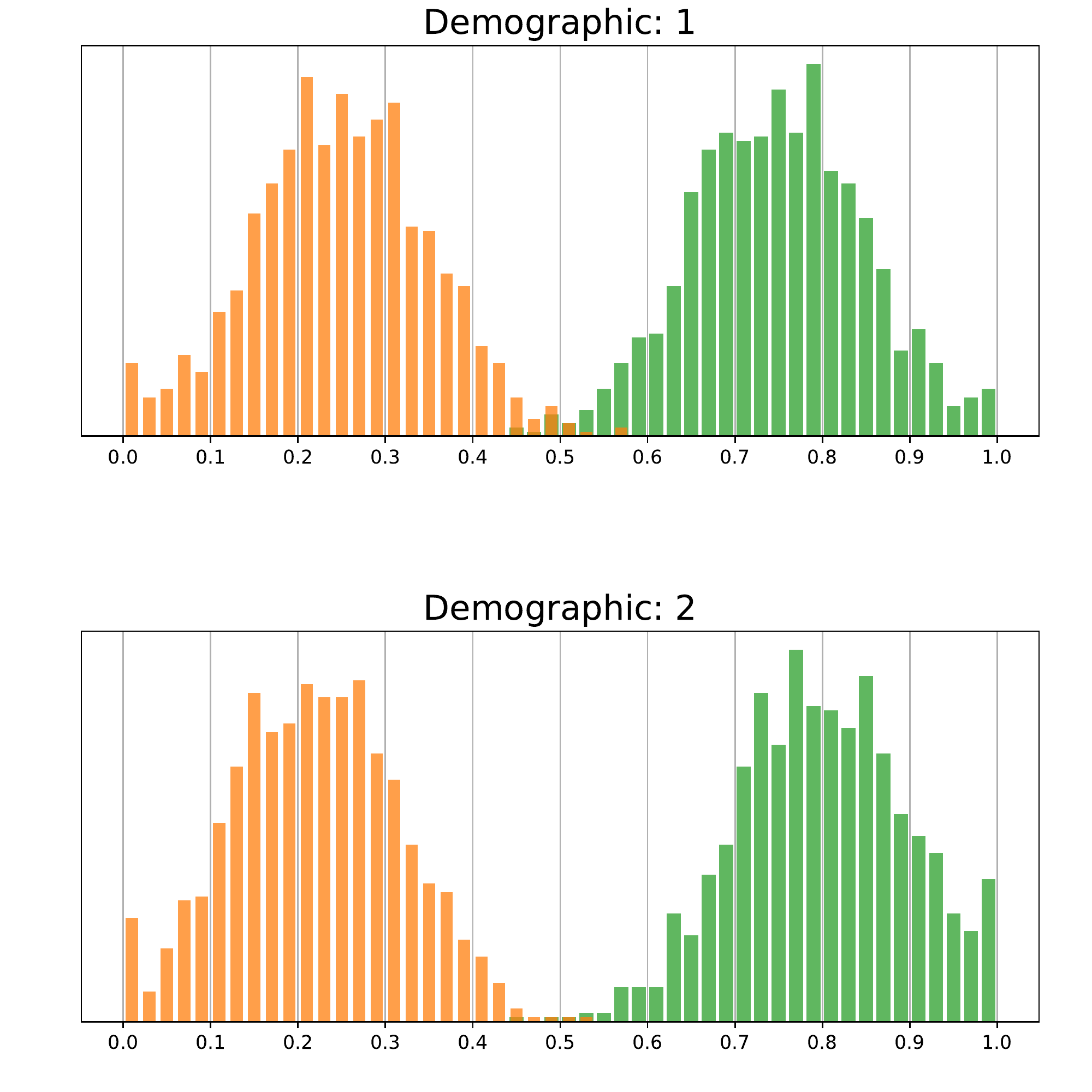}
    \caption{Fair BV system}
    \label{fig:fixed_fair}
\end{subfigure}
\quad \quad %
\begin{subfigure}[b]{0.4\textwidth}
    \includegraphics[width=\textwidth]{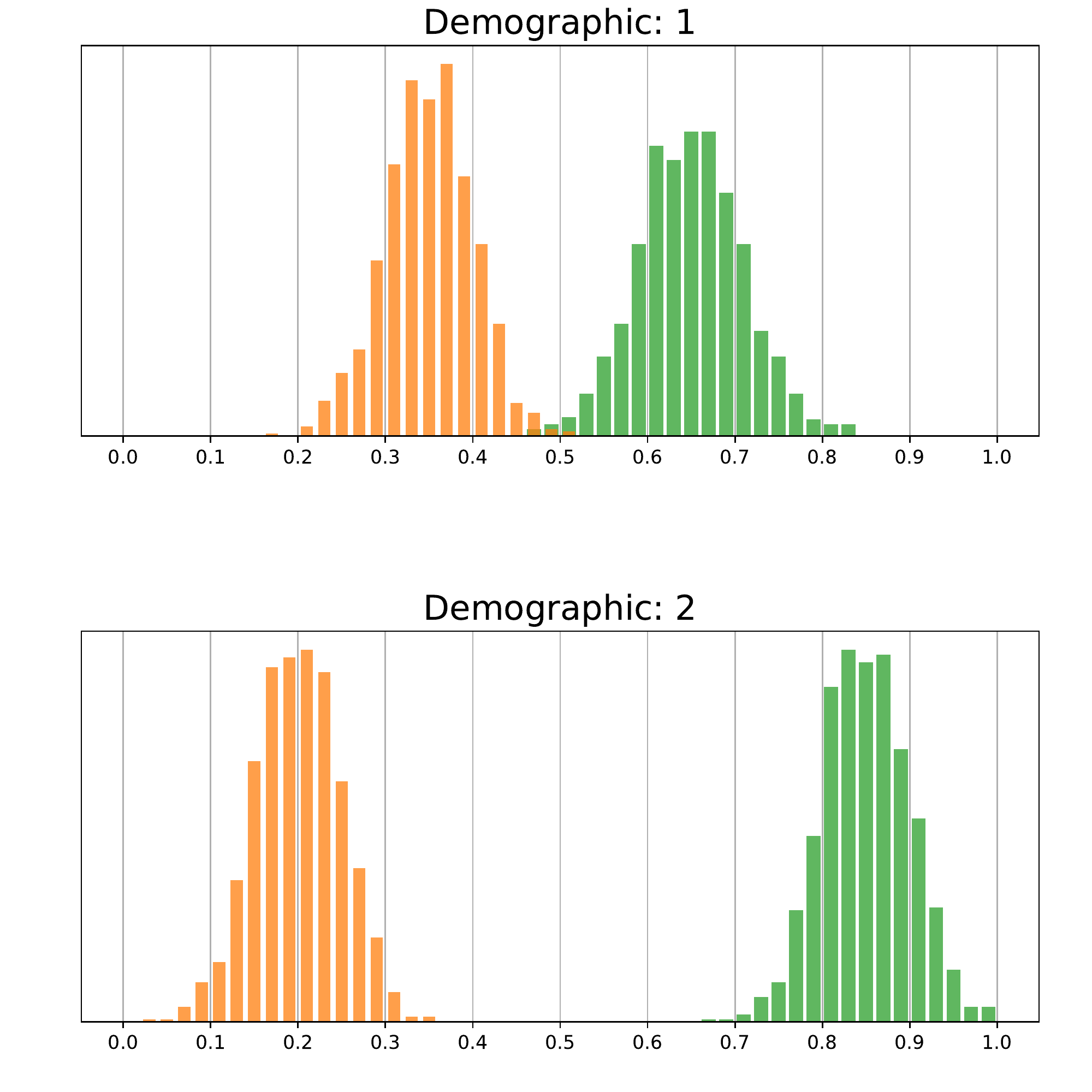}
    \caption{Unfair BV system}
    \label{fig:fixed_unfair}
\end{subfigure}
\caption{Score distributions of canonical biometric verification systems with
        two demographic groups ($d_1$ and $d_2$). Distributions of genuine and
        imposter scores are represented by green and orange bars, respectively.
        The systems in (a) and (b) exhibit similar overall verification
        accuracies. However, their score distributions reveal demographic
        disparity.}
\label{fig:fixed_example}
\end{figure}

Majority of the existing work in evaluating the fairness or demographic
equitability of the biometric system is based on the differential outcome. These
measures are easy to calculate, based on well-established error rates, and treat
the biometric system as a complete black box. The assessment of demographic
fairness of biometric system based on differential performance has received
little attention. The use of distributions of genuine and imposter scores of a
biometric recognition system towards evaluation of its demographic fairness has
several advantages: first, the measures based on differential outcome evaluate
fairness of the system entirely from the (number of) samples causing incorrect
decisions. While these extremal samples signify the accuracy of the biometric
system, we believe that the demographic fairness needs to be evaluated across
all samples, irrespective of their recognition outcome. The use of differential
performance (score distributions) facilitates consideration of entire sample set
towards fairness assessment. Second, the incorrect decisions (characterized by
FMR and FNMR) are dependent on a score threshold. This variable is either fixed
or computed from probably different set of data. In the former case, one has to
evaluate fairness at multiple score thresholds to interpret the equitability of
the biometric system. In the latter case, the score threshold is sensitive to, and
hence, impacted by the distribution of unseen, disjoint set of data (and
underlying demographic)~\cite{hupont_demogpairs}. The measures based on score
distributions do not involve such threshold, and thus, can be computed without
interference of such external parameters. Third, the score threshold, being
external, is often easy to \textit{tune} as per the need of the application. The
fairness measures based on differential performance are agnostic to such tuning,
and represent the fairness of underlying core mechanism of the biometric system.

Consider two (canonical) biometric verification systems---whose score
distributions are shown in
Figs.~\ref{fig:fixed_fair}--\ref{fig:fixed_unfair}. For both systems, the
classification accuracy (in terms of FMR and FNMR) is nearly the same. However, the
first system (Fig.~\ref{fig:fixed_fair}) is likely to be fair to both
demographic groups (top and bottom row), whereas the other biometric system is
likely to be unfair to demographic $d_1$ (top row). We believe that such
disparity among demographic groups, beyond the recognition accuracy, needs to be
quantified systematically. In this work, we propose three measures based
on differential performance for evaluation of demographic fairness of a
biometric verification system. The fundamental component of each measure is
calculated on distributions of an individual demographic group. Depending on
fusion of these components to obtain a final measure, we define three different
variants of each measure. We also provide a solution to reduce the effect of
demographic imbalance in the test dataset towards fairness evaluation.  We
explain the behavior of each fairness measure followed by illustration on
canonical data (synthetic, yet realistic). Our contributions can be summarized
as follows:

\begin{itemize}
\item We propose three measures for evaluation of demographic fairness of
biometric verification systems.\footnote{A Python implementation of each
evaluation measure is provided at: \url{https://gitlab.idiap.ch/bob/bob.paper.icpr2022_fairness_index_measures2022}.}
Our measures, being based on scores,
consider how well a pair of samples (genuine or imposter) has been matched,
rather than just `whether it has been matched'.

\item We propose a weighted fusion strategy to account for demographic imbalance
in the benchmarking datasets. Our weighing strategy attempts to provide higher
importance to relatively under-represented demographic groups.

\item We propose three different variants for each fairness measure to facilitate
assessment of fairness from multiple perspectives. 

\item Being agnostic to the modality, the proposed measures are applicable
across various biometric modalities.
\end{itemize}

In Section~\ref{sec:nfm}, we formulate the problem of algorithmic fairness in
general and biometric verification-specific context. The weighted-fusion
strategy and fairness measures are proposed in Section~\ref{sec:measures}.
Summary is presented in Section~\ref{sec:summary}.



\section{Problem Formulation}
\label{sec:nfm}
We begin with general definition of algorithmic fairness in biometrics followed
by discussion on how the notion of fairness applies to the problem of biometric
verification (BV). 

\subsection{Fairness in Biometrics}
\label{ss:fv_general}

In~\cite{fairness_define}, Mehrabi~\textit{et al.} provide several definitions
for algorithmic fairness in machine learning. The definition related to
demographic parity (Def.~3, Sec.~4.1), which is more suitable for the present
discussion, suggests that the likelihood of the positive outcome should be the
same regardless of whether the person is in the protected group. For a
biometric system, the term protected group may be used to refer to different
demographics present in data. The demographic division could be based on the
factors such as gender, race, or ethnicity. While these factors are often
regarded as sensitive issues, the authors discuss these attributes of data
purely from technical aspect, and provide solutions (\textit{i.e.} fairness
measures) that are generic towards any multi-demographic attribute. In this
paper, the term `demographic' refers to subset(s) of data---where the
partitioning is possible due to any demographic attribute.

Let $\mathcal{T}$ be the test dataset, consisting of $N$ samples (biometric
presentations), used to benchmark the fairness of a given biometric system.  We
assume that the presentations are correctly acquired, labelled, and processed
to obtain feature descriptors.  Let the dataset consist of samples from $K$
demographic groups, such that $\mathcal{D} = \{d_1, d_2, \cdots, d_K \}$. Here,
$K > 1$ is a finite, but usually small number (For $K=1$, the dataset
represents homogeneity in terms of demographic variation).  Additionally, we
assume that each sample can be associated with one and only one demographic
from $\mathcal{D}$. That is, demographic subsets $d_1, d_2, \cdots, d_K$ are
disjoint. This assumption may not always hold true in some real-world scenarios
due to finite number of demographic classes, and (quite often) subjective
assignment of these classes to the samples. Notwithstanding with these
imperfections, we consider samples in $\mathcal{T}$ to be disjoint in terms of
demographic categorization; and therefore, if a demographic $d_i, \, i = 1, 2,
\cdots, K$ consists of $N_i$ samples ($N_i > 0$), we have $ \sum_{i=1}^{K} N_i
= N$. A biometric system, with predictor function $\mathtt{F}$, is considered
to be fair, if $P(\mathtt{F} | d_i) = P(\mathtt{F} | d_j)$  for every pair of
$d_i, \, d_j \in \mathcal{D}$~\cite{fairness_define,fdr2021}. Thus, parameters
of distribution characterizing different demographic groups need to be analyzed
for fairness of the system.

%
\subsection{Fairness in Biometric Verification}
\label{ss:fv_problem}

In this work, we discuss fairness for a BV system
which is one of the most important and widely deployed biometric application.
The BV system is a 1:1 matching process that compares the features of probe
sample with the features of previously enrolled sample of the claimed identity.
When both samples belong to the same identity, the match is considered to be
\textit{genuine} and a match between features of two different identities is
regarded as \textit{imposter}. A score threshold ($\tau$), fixed or computed
over different dataset, binarizes the matching scores into decisions (match or
no-match). If $\mathbf{e}$ and $\mathbf{p}$ represent the feature vectors
(generated by the BV system) of enrolled and probe samples, respectively; a
matching score is calculated using a score function $s_f(\mathbf{e},
\mathbf{p})$. Most often, the score function comprises Euclidean or cosine
distances followed by suitable modifications to ensure that the matching score
is a real-valued scalar in [0, 1] with higher values indicating a better match.
For an ideal BV system, $s_{f}(\mathbf{e}, \mathbf{p}) \to 1$ for a genuine
match, and $s_{f}(\mathbf{e}, \mathbf{p}) \to 0$ for an imposter pair. 

For a BV system, the algorithmic fairness can be defined as the ability of the
BV system to exhibit similar performance-- in terms of scores, accuracies, or
error rates-- to different demographic groups in the test data. However,
quantitative measurement of the said term is necessary in order to
systematically assess the fairness of BV systems. In our work, the fairness
test is performed by obtaining the feature vectors or \textit{embeddings} for
$N$ pairs from $\mathcal{T}$. The matching scores are obtained for multiple
pairs of ($\mathbf{e}, \mathbf{p}$) samples by computing score function $s_f$.
For a genuine match, the features $\mathbf{e}$ and $\mathbf{p}$ refer to the
same subject, and hence, both samples essentially belong to the same
demographic in $\mathcal{D}$. For an imposter match, on the other hand, both
identities may or may not belong to the same demographic in $\mathcal{D}$. In
this work, however, the imposter matches (and thus, corresponding score
distributions) are restricted to the pair of samples from different identities,
yet belonging to the same demographic. This experimental setup, thus, evaluates
or defines the imposter matches in \textit{intra-demographic} manner. One
rationale behind this setup is to analyze the behavior of the BV system,
towards each demographic independently. Also, it has been demonstrated that the
number of false matches from inter-demographic pairs is much smaller than those
from the same demographic~\cite{fr_children2018,nist3,howard2019}. Therefore,
for fairness evaluation, we restrict the choice of sample pairs selected within
same demographic group.



\section{Proposed Fairness Measures}
\label{sec:measures}
%

Prior to the introduction of fairness measures, we discuss strategy to reduce
the impact of demographic imbalance in test data. 

\subsection{Fusion for Demographic Imbalance in Test Datasets}

The fairness of a BV system refers to its equitability across different
demographic groups. The test or benchmarking dataset acts as a proxy to
real-world scenario with a fundamental assumption that these data represent a
reasonably similar distribution of real-world data w.r.t. multiple attributes.
However, studies have noted that many biometric datasets are highly imbalanced
in distributions of their demographic
attributes~\cite{garcia2019harms,hupont_demogpairs}. To the best of our
knowledge, existing evaluations of fairness do not take into account the fact
that, in most cases, the test datasets under-represent some demographic groups.
The disparity in sample sizes in the training set is beyond the scope of this
work. However, the question: ``should the imbalance in sample sizes of
different demographic groups be considered during evaluation of fairness?''
needs to be addressed. Such a consideration can be helpful while comparing the
fairness of a system across datasets (or in dataset-agnostic manner). An
intuitive solution would be designing a weighing strategy (weighted-fusion
rule) where a weight relative to the cardinality (sample size) of each
demographic group can be assigned while computing overall fairness measure. If
$w_i$ is the weight assigned to the demographic group $d_i$, then $w_i = f
(N_i), \, i = 1, 2, \cdots, K$.

For several use-cases, the fusion weights are monotonically non-decreasing
functions of (or directly proportional to) the sample size, \textit{i.e.}, $w_i
\propto N_i$. The normalized weights are related to the probability of a sample
drawn from the test dataset $\mathcal{T}$ belonging to $d_i$. With this
approach, the fairness measure of under-represented demographic groups (where
few $N_i$'s are significantly smaller than remaining ones) are assigned smaller
weights during the fusion (weighted linear combination). The imbalance in
demographic data (training or testing) has adverse impact on the fairness of
the biometric system \cite{hupont_demogpairs}. Unfortunately, the
aforementioned approach of computing fusion weights suppresses the measures for
those demographic groups---for which the BV system is likely to be unfair.
Alternatively, the fusion weights could be inversely proportional to (or
monotonically non-increasing functions of) the sample size of the demographic
($w_i \propto \frac{1}{N_i}$). This weighing approach emphasizes the
contributions of under-represented demographic groups. While this approach
appears to be reasonable for minor variations in sample sizes, it is not robust
for highly unbalanced test datasets. If the proportion of the sample size of
any demographic group is tiny $\left(\frac{N_i}{N} \to 0 \right)$, the
corresponding measure will be assigned excessively high fusion-weight. With
only few test samples, it is indeed difficult to ascertain whether these
samples represent true distribution of the underlying demographic group.
Therefore, assigning high weights to such groups may not reflect true fairness
of the BV system, either. We propose the following mechanism to compute fusion
weights, towards obtaining final fairness measure, for demographic groups in
$\mathcal{D}$.

\begin{figure}[!t]
\centering
\includegraphics[width=0.4\textwidth]{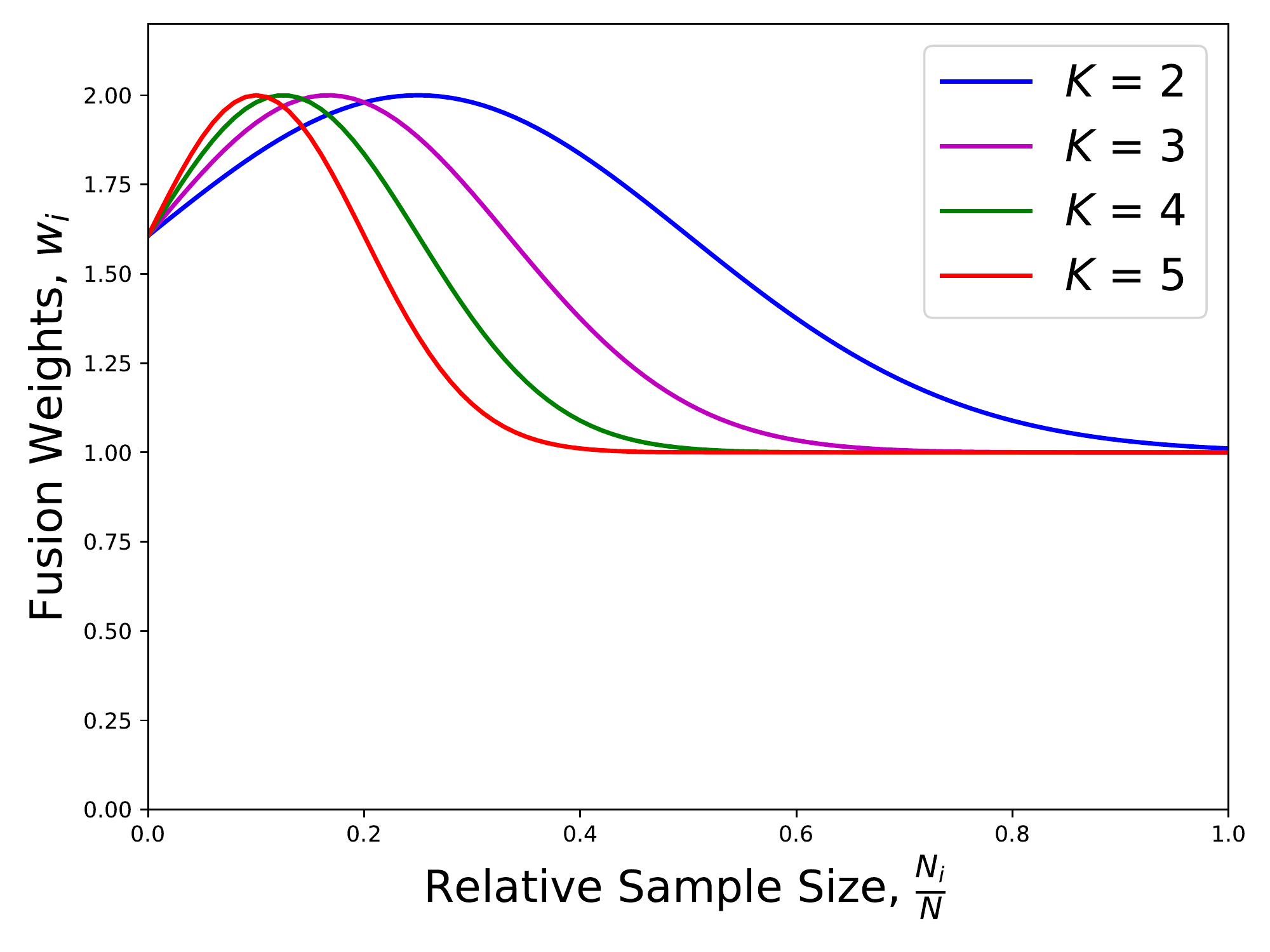}
\caption{Fusion weights (without normalization) as a function of relative 
sample size illustrated for different number of demographic groups ($K$).}
\label{fig:w_plots}
\vspace{-4mm}
\end{figure}
%
For a test dataset comprising $K$-demographic groups, the ideal data
distribution refers to (nearly) equal sample size of each demographic group
(\textit{i.e.}, $N_i \approx N/K, \,\, \forall \, i$). A demographic with
relatively minor representation than the ideal one should be assigned with
higher fusion weights to provide higher attention to its fairness score.
However, if the sample size of a particular group is substantially small, the
fusion weight should be nominally controlled to avoid a possibility of
insufficient samples affecting the fairness score of the BV system.
Consequently, we propose a weighing function where the fusion weights are
calculated through exponential function of the sample size $N_i$.  The
parameters of the exponential are determined from the properties of the test
dataset. Equation~\ref{eq:w_formula} provides the formula for fusion weights.
\begin{align}
\label{eq:w_formula}    
\hat{w_i} &= c + \exp {\left(\frac{-1}{2 \sigma^2} \left(\frac{N_i}{N} - \frac{1}{2 K} \right)^2 \right)},  \\
w_i &= \frac{\hat{w_i}} {\sum_i \hat{w_i}} \quad i = 1, 2, \cdots K;
\end{align}
where, $\sigma = \frac{1}{2 K}$ represents the standard deviation of the
weighing function. The numeric constant, $c$ set to 1.0, provides numerical
stability to the weight calculation, and also limits drastic variations across
fusion weights. The unnormalized and normalized fusion weights are represented
by $\hat{w}$ and $w$, respectively. Figure~(\ref{fig:w_plots}) shows the
unnormalized fusion weights for $K = 2, 3, 4, 5$. It may be observed that
fusion weights (\textit{i.e.} relative importance) are assigned higher values
for marginally under-represented demographic groups; however, further reduction
in sample size does not encourage similar importance (due to possibly
inaccurate representation of underlying demographic data). On the other hand,
the over-represented demographic group is assigned relatively lesser importance
to avoid its dominance on determining the fairness of the BV system. 

\subsection{Fairness Measures}

A primary component (akin to the building block) of a particular fairness
measure is first calculated separately for each demographic group in
$\mathcal{D}$. The fairness measure for overall test dataset $\mathcal{T}$ is
obtained by combining the corresponding resultants (of primary component) for
each demographic group. In each case, the output of the scoring function is
normalized in the range [0, 1]. The values of fairness measures are dependent
on the chosen score function (\textit{i.e.}, distance function).\\

\noindent
\textbf{1. Separation Fairness Index (SFI):} 

\noindent We define the Separation Fairness Index (SFI) of the biometric
verification system as the measure of its equitability towards separation of
expected values of genuine and imposter scores across constituent demographic
groups. 

\begin{figure}[t]
\centering
\begin{subfigure}[b]{0.3\textwidth}
    \includegraphics[width=\textwidth]{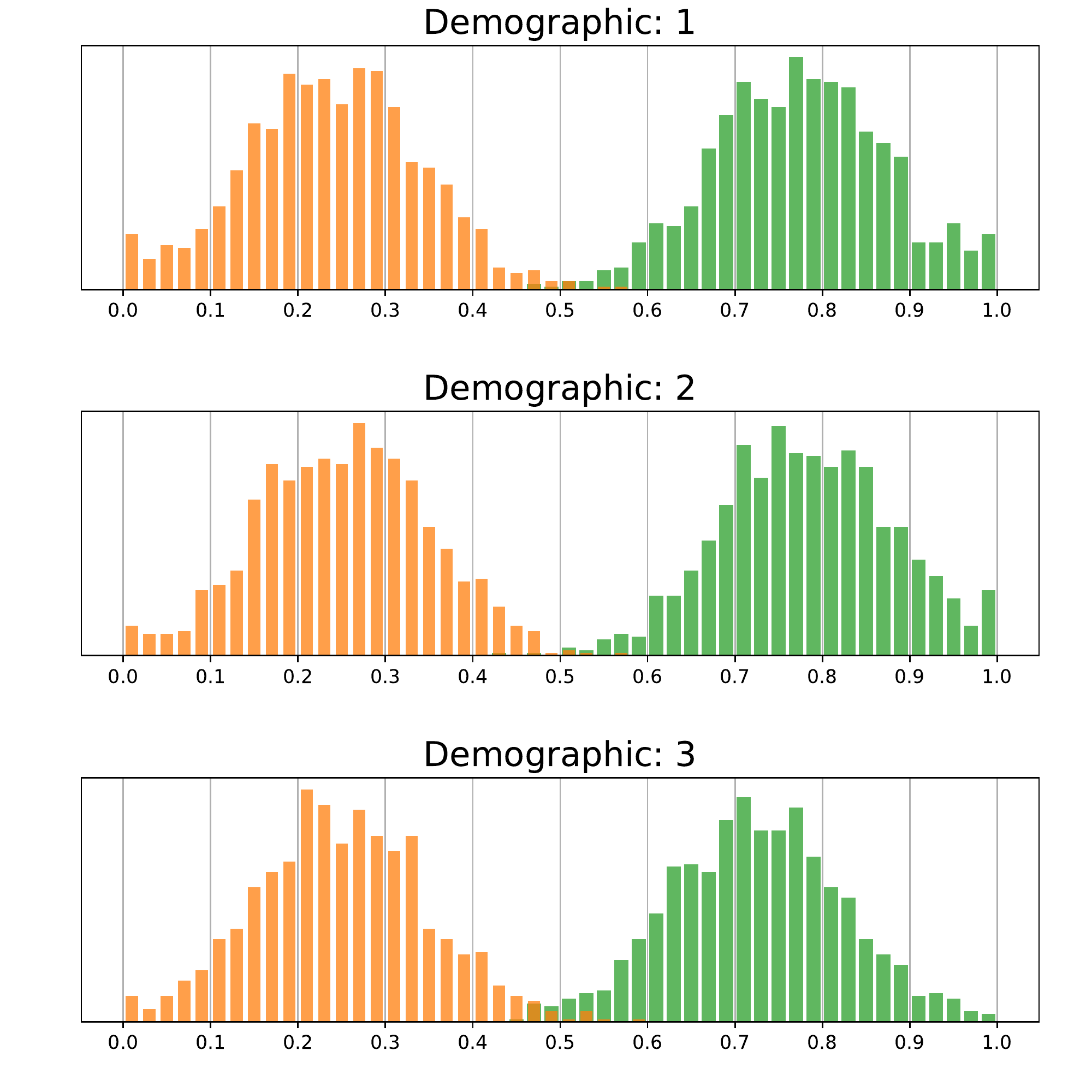}
    \caption{Fair BV}
    \label{fig:sfi_fair}
\end{subfigure}
\,
\begin{subfigure}[b]{0.3\textwidth}
    \includegraphics[width=\textwidth]{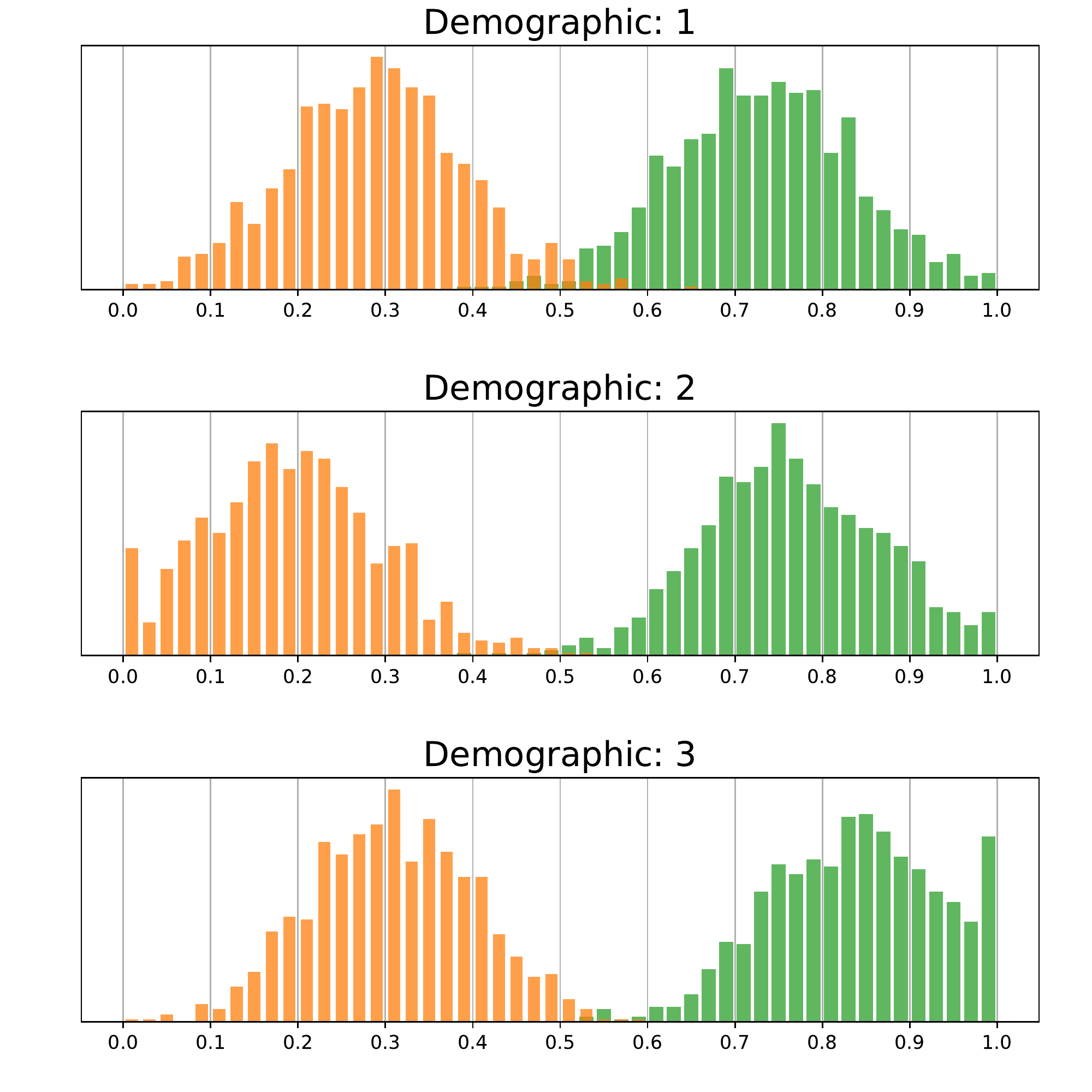}
    \caption{Unfair BV}
    \label{fig:sfi_unfair}
\end{subfigure}
\,
\begin{subfigure}[b]{0.3\textwidth}
    \includegraphics[width=\textwidth]{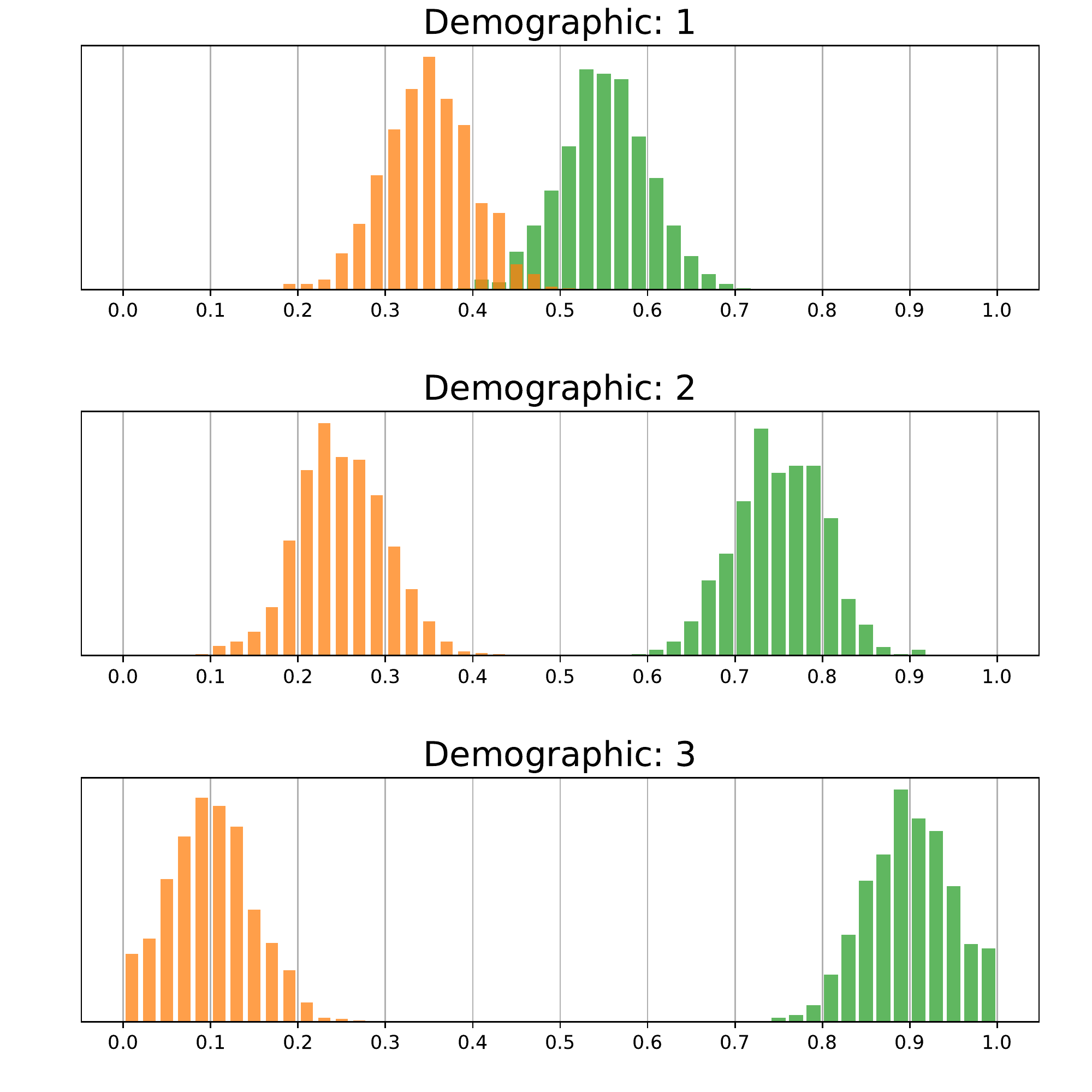}
    \caption{Highly Unfair BV}
    \label{fig:sfi_extreme}
\end{subfigure}
\caption{Score distributions of a canonical biometric verification systems with
        three demographic groups evaluated for SFI measure. Histograms of
        genuine and imposter scores are represented by green and orange bars,
        respectively.}
\label{fig:sfi_n}
\end{figure}
%
Consider a BV system that generates feature vectors for samples of demographic
group $d_i$. The matching scores for pairs of the feature vectors $\mathbf{e}$
and $\mathbf{p}$ are calculated using score function, $s_f$. Let $\mu_G$ be the
expected value of matching scores obtained for genuine pairs ($\mathbf{e}$ and
$\mathbf{p}$ belong to the same subject), and $\mu_I$ be the expected value of
imposter pairs ($\mathbf{e}$ and $\mathbf{p}$ belong to different subjects, but
within same demographic). Higher separation between $\mu_G$ and $\mu_I$
generally leads to better verification rates. If we denote the separation
between two expected values for demographic group $d_i$ as $z_{S_i}$, a fair BV
system is expected to exhibit similar values of $z_{S_i}$ for each demographic
group in $\mathcal{D}$. We define the normal (or absolute) variant of SFI as
the following.
\begin{align}
\label{eq:sfi_n}    
&z_{S_i} = | \mu_{G_i} - \mu_{I_i} |,  \quad i = 1, 2, \cdots, K \nonumber \\
&z_{S_\mathrm{mean}} = \frac{1}{K} \sum_{i=1}^{K} z_{S_i} \nonumber \\
&\mathrm{SFI_N} = 1 - \frac{2}{K} \sum_{i=1}^{K} |z_{S_i} - z_{S_\mathrm{mean}} |.
\end{align}

For a fair BV system, the $\mathrm{SFI_N}$ reaches a value of 1.0 indicating
similar (and hence, fair) separation of expected values of genuine and imposter
scores across demographic groups. If the BV system exhibits unequal separation
of such scores, the FSI value decreases accordingly with the worst-case value
being 0. Fig.~\ref{fig:sfi_n} depicts these scenarios on a canonical
(synthetic) data.

Sometimes the performance of overall system is characterized by the worst
performing sub-system. Depending on the nature of performance measure, the
minimum or maximum fusion rule is applied. Accordingly, we suggest an extremal
variant of the FSI measure where maximum value of discrepancy in separation is
chosen to represent the fairness of the BV system. The extremal variant of SFI,
denoted as $\mathrm{SFI_E}$ is provided by Eq.~\ref{eq:sfi_e}.
\begin{align}
\label{eq:sfi_e}    
\mathrm{SFI_E} = 1 - 2 \, \max_i \, \big|z_{S_i} - z_{S_\mathrm{mean}} \big|,
\end{align}
where intermittent variables computed as described previously in
Eq.~\ref{eq:sfi_n}. The normal and extremal variants of SFI do not consider any
variations in the sample sizes of demographic groups. We propose a weighted
variant of the SFI using the weighing strategy discussed earlier in this
section (Eq.~\ref{eq:w_formula}), and re-using the calculation from
Eq.~\ref{eq:sfi_e} for group-level variables. The resultant fairness
measure is given by Eq.~\ref{eq:sfi_w}.
\begin{align}
\label{eq:sfi_w}    
\mathrm{SFI_W} = 1 - 2 \, \sum_{i=1}^{K} w_i \, (|z_{S_i} - z_{S_\mathrm{mean}} |).
\end{align}

Being a linear combination with normalized weights ($\sum_i w_i = 1$), the
overall range and behavior of SFI\textsubscript{W} is similar to that of normal
variant of SFI, however, it \textit{scales} the combination based on sample
sizes of demographic groups. Note that a highly accurate BV system may still be
considered less fair, if the separation across groups is not uniform. For the
canonical BV systems, whose score distributions are depicted in
Fig.~\ref{fig:sfi_n}, the FSI values are provided in Table~\ref{tb:sfi_n}.\\

\begin{table}[b]
\renewcommand{\arraystretch}{1.0}
\centering
\caption{The values of the SFI measure on canonical BV systems. For balanced
        case, 1000 samples were selected for each demographic group. For
        imbalanced case, the number of samples per demographic group was
        modified to [100, 1000, 2000].}
\label{tb:sfi_n}
\begin{tabular}{ l | p{1.5cm} p{1.5cm} p{1.5cm} || p{1.5cm} p{1.5cm} p{1.5cm}} \hline			
 & \multicolumn{3}{c ||}{Balanced} &\multicolumn{3}{c}{Imbalanced} \\ \cline{2-7}
 & Fair & Unfair & Highly Unfair & Fair & Unfair & Highly Unfair \\ \hline
$\mathrm{SFI_N}$ & 0.9489 & 0.9064 & 0.6017 &  0.9833 & 0.8424 & 0.5975 \\
$\mathrm{SFI_E}$ & 0.9233 & 0.8596 & 0.4025  & 0.9750 & 0.7636 & 0.3963 \\
$\mathrm{SFI_W}$ & 0.9489 & 0.9064 & 0.6017 &  0.9846 & 0.8292  & 0.6230 \\ \hline
\end{tabular}
\end{table}


\noindent\textbf{2. Compactness Fairness Index (CFI):}

\noindent We define the Compactness Fairness Index (CFI) of the biometric
verification system as the measure of its equitability towards compactness (or
spread) of genuine scores and imposter scores across constituent demographic
groups. 

For a BV system, let $\sigma_G$ and $\sigma_I$ be the standard deviations of
the matching scores obtained for genuine and imposter pairs, respectively.
Smaller values of spread (measured in terms of standard deviation here) for
both genuine and imposter scores are desirable, though this characteristic
alone does not determine accuracy of verification. Consistency of spread of
score distributions across demographic groups simplifies several scaling and
normalization procedures-- even if applied using global statistics of
$\mathcal{T}$.

\begin{figure}[!t]
\centering
\begin{subfigure}[b]{0.3\textwidth}
    \includegraphics[width=\textwidth]{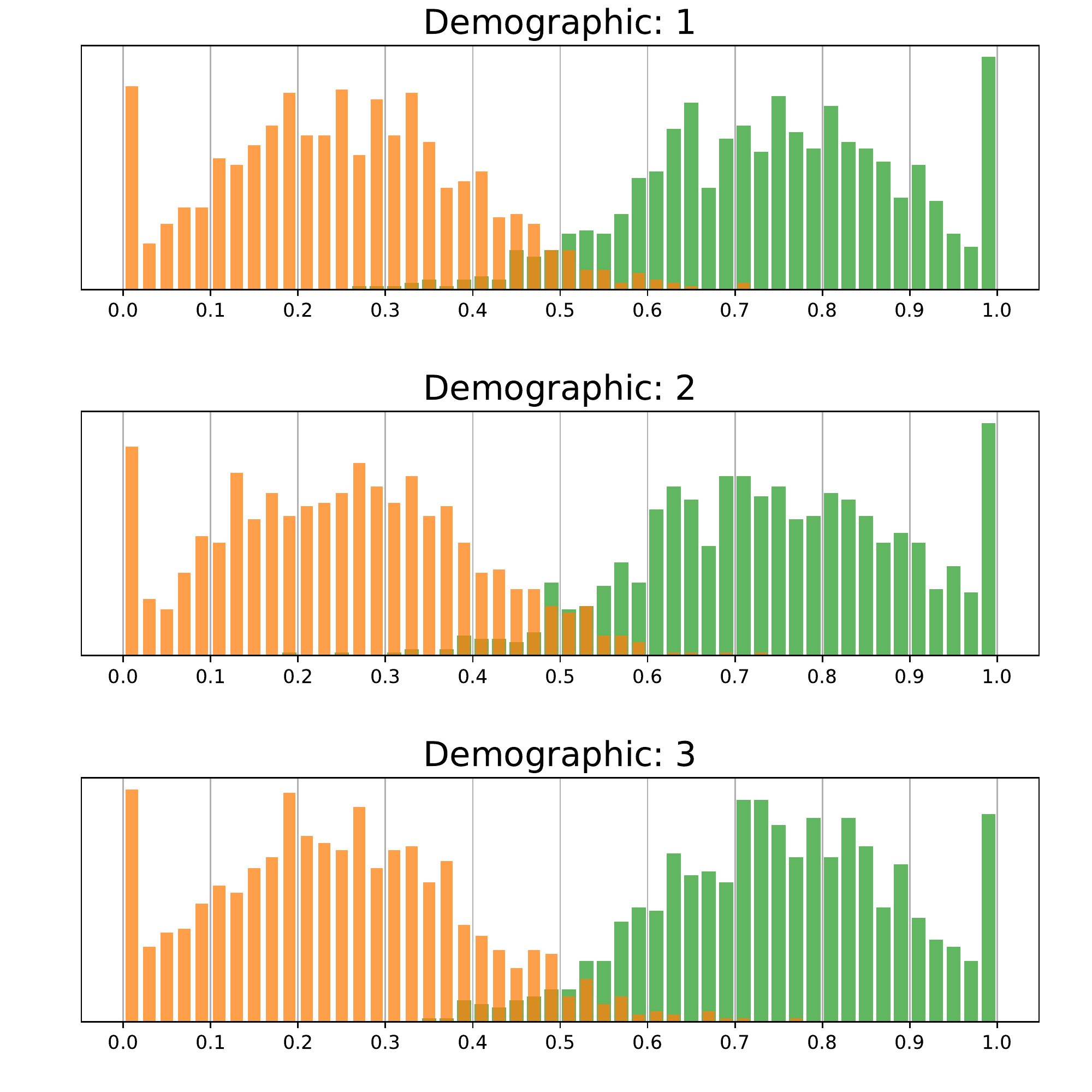}
    \caption{Fair BV}
    \label{fig:cfi_fair}
\end{subfigure}
\, %
\begin{subfigure}[b]{0.3\textwidth}
    \includegraphics[width=\textwidth]{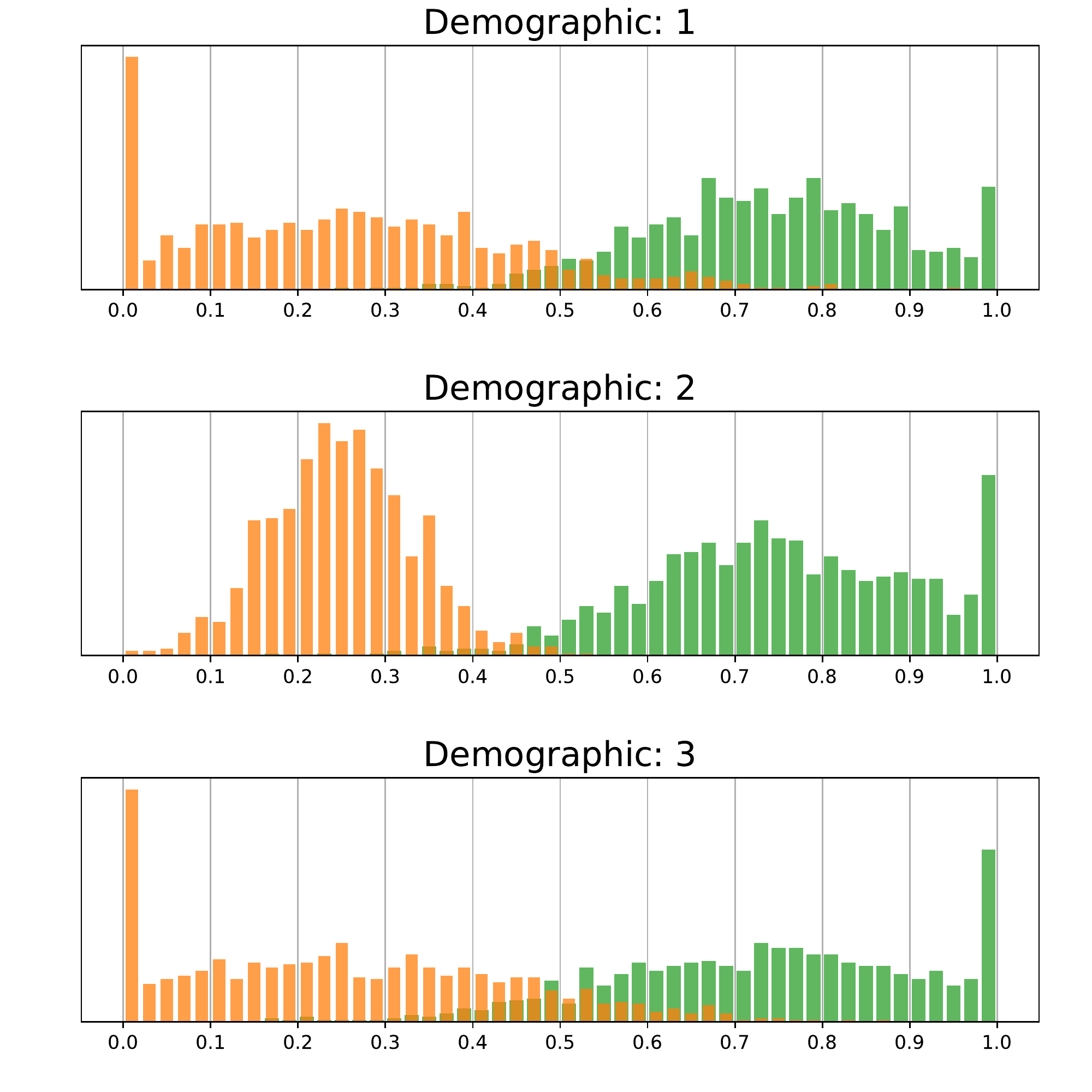}
    \caption{Unfair BV}
    \label{fig:cfi_unfair}
\end{subfigure}
\,
\begin{subfigure}[b]{0.3\textwidth}
    \includegraphics[width=\textwidth]{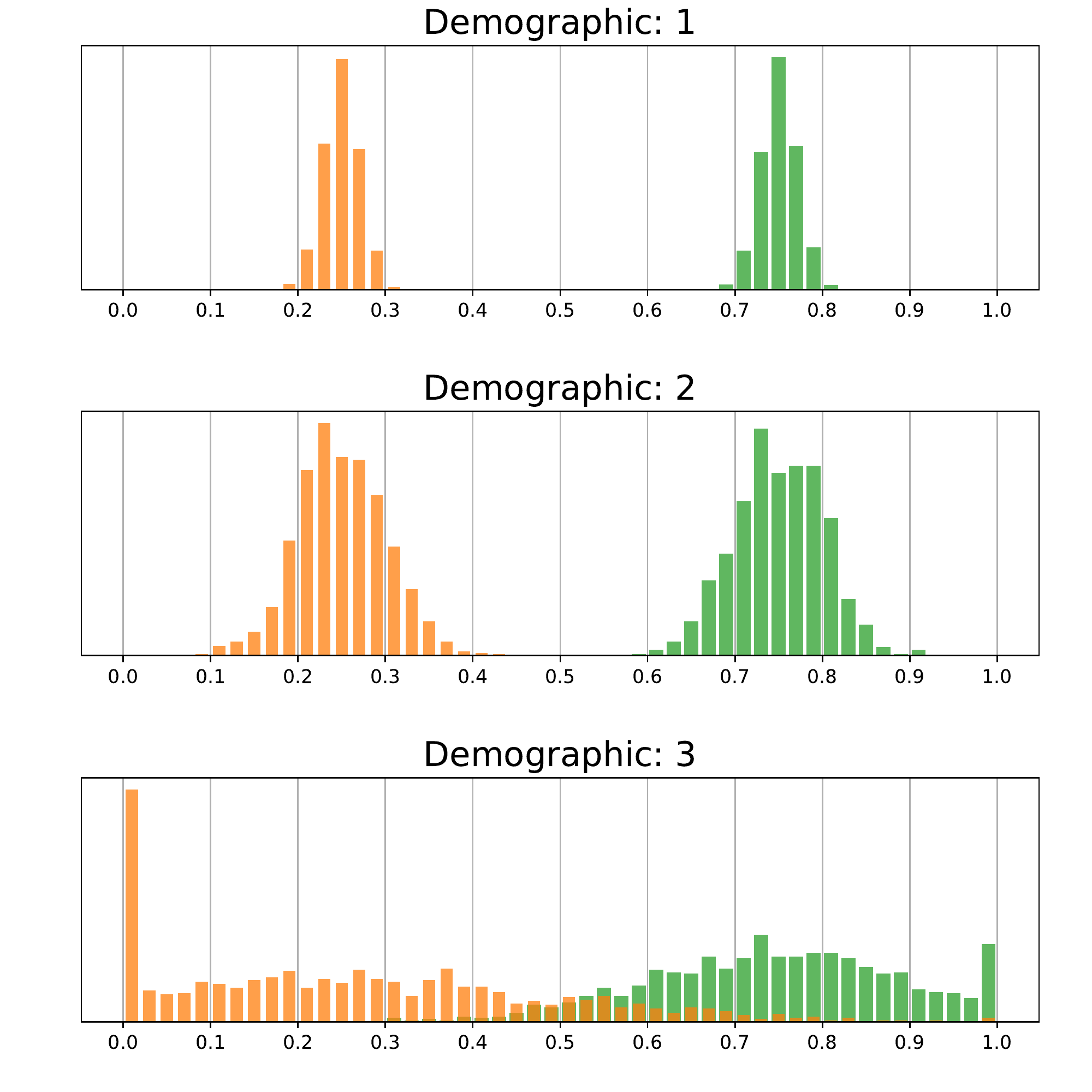}
    \caption{Highly Unfair BV}
    \label{fig:cfi_extreme}
\end{subfigure}
\caption{Score distributions of a canonical biometric verification systems with
        three demographic groups evaluated for CFI measure. Histogram of
        genuine and imposter scores are represented by green and orange bars,
        respectively.}
\label{fig:cfi_n}
\end{figure}

If we denote the combined spread of both score distributions for demographic
group $d_i$ as $z_{C_i}$, a fair BV system is expected to exhibit similar
values of $z_{C_i}$ for each demographic group in $\mathcal{D}$. We define the
normal (or absolute) variant of CFI as the following.
\begin{align}
\label{eq:cfi_n}    
&z_{C_i} = \sigma_{G_i} + \sigma_{I_i},  \quad i = 1, 2, \cdots, K \nonumber \\
&z_{C_\mathrm{mean}} = \frac{1}{K} \, \sum_{i=1}^{K} z_{C_i} \nonumber \\
&\mathrm{CFI_N} = 1 - \frac{2}{K} \, \sum_{i=1}^{K} (|z_{C_i} - z_{C_\mathrm{mean}}|).
\end{align}

\begin{table}[!b]
\renewcommand{\arraystretch}{1.0}
\centering
\caption{The values of the CFI measure on canonical BV systems. For balanced
        case, 1000 samples were selected for each demographic group. For
        imbalanced case, the number of samples per demographic group was
        modified to [100, 1000, 2000].}
\label{tb:cfi_n}
\begin{tabular}{ l | p{1.5cm} p{1.5cm} p{1.5cm} || p{1.5cm} p{1.5cm} p{1.5cm}} \hline			
    & \multicolumn{3}{c||}{Balanced} &\multicolumn{3}{c}{Imbalanced} \\ \cline{2-7}
    & Fair & Unfair & Highly Unfair & Fair & Unfair & Highly Unfair \\ \hline
$\mathrm{CFI_N}$ & 0.9930 & 0.9157 & 0.7921 & 0.9888 & 0.9195 &  0.7901\\
$\mathrm{CFI_E}$ & 0.9895 & 0.8735 & 0.6882 & 0.9832 & 0.8793 &  0.6851\\
$\mathrm{CFI_W}$ & 0.9930 & 0.9157 & 0.7921 & 0.9893 & 0.9191 &  0.8052\\ \hline
\end{tabular}
\end{table}

Similar to SFI, the best possible value for $\mathrm{CFI_N}$ is 1.0---where the
combined spread of genuine and imposter score distributions for each
demographic group is the same. With disparity in the combined spread of scores
across demographics, the $\mathrm{CFI_N}$ value decreases, indicating that for
some groups the BV system does not generate enough compact representations in
feature space than others. The behavior of CFI\textsubscript{N} is graphically
depicted in Fig.~\ref{fig:cfi_n}. 
An extremal variant of the CFI measure, as given in Eq.~\ref{eq:cfi_e}, is
based on the maximum discrepancy in the spread of scores for a particular
demographic group in $\mathcal{D}$. 
\begin{align}
\label{eq:cfi_e}    
\mathrm{CFI_E} = 1 - 2 \, \max_i (|z_{C_i} - z_{C_\mathrm{mean}} |).
\end{align}

The weighted variant of the CFI weighs the spread variables ($z_{C_i}$)
depending on the sample size of the demographic group, $d_i$. The fusion weights
are calculated as discussed earlier in Eq.~\ref{eq:w_formula}. The expression
for CFI\textsubscript{W} is given below in Eq.~\ref{eq:cfi_w}.
Table~\ref{tb:cfi_n} provides the values of all variants of CFI computed on
balanced as well as imbalanced canonical datasets.
\begin{align}
\label{eq:cfi_w}    
\mathrm{CFI_W} = 1 - 2 \, \sum_{i=1}^{K} w_i \, (|z_{C_i} - z_{C_\mathrm{mean}} |).
\end{align}


\noindent\textbf{3. Distribution Fairness Index (DFI):}

\noindent We define the Distribution Fairness Index (DFI) of the biometric
verification system as the measure of its equitability towards overall score
distributions across constituent demographic groups. 

\begin{figure}[!t]
\centering
\begin{subfigure}[b]{0.3\textwidth}
    \includegraphics[width=\textwidth]{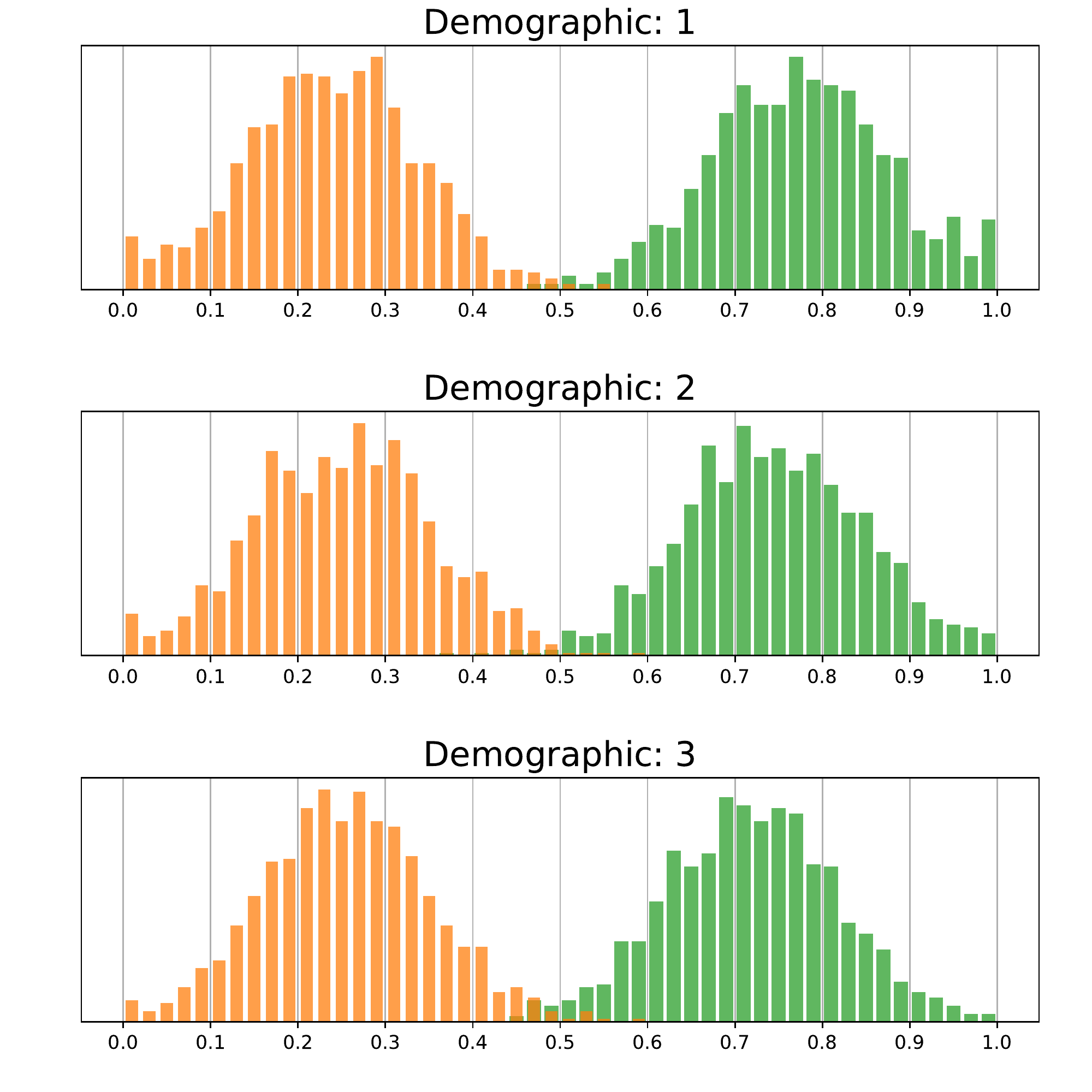}
    \caption{Fair BV}
    \label{fig:dfi_fair}
\end{subfigure}
\,%
\begin{subfigure}[b]{0.3\textwidth}
    \includegraphics[width=\textwidth]{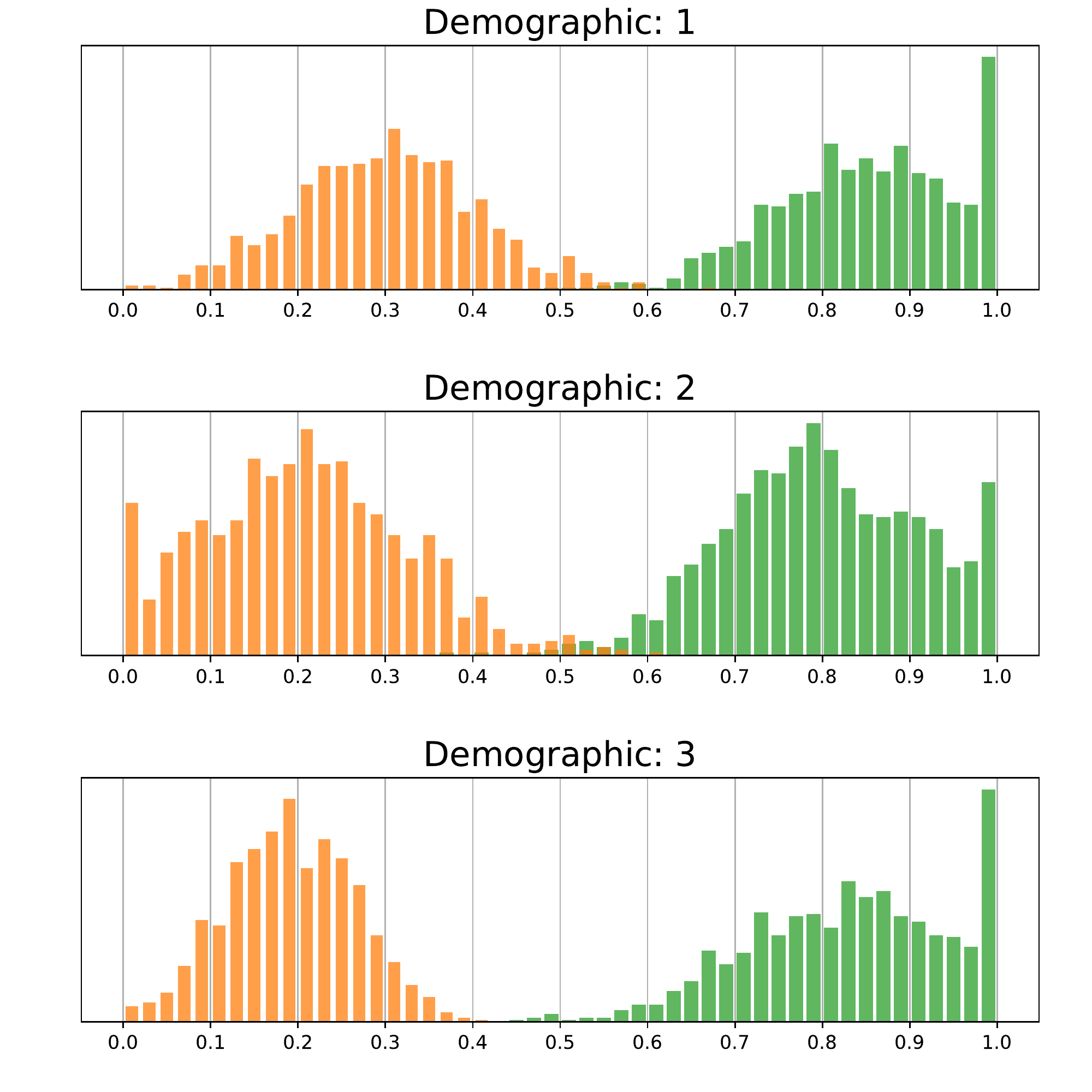}
    \caption{Unfair BV}
    \label{fig:dfi_unfair}
\end{subfigure}
\,
\begin{subfigure}[b]{0.3\textwidth}
    \includegraphics[width=\textwidth]{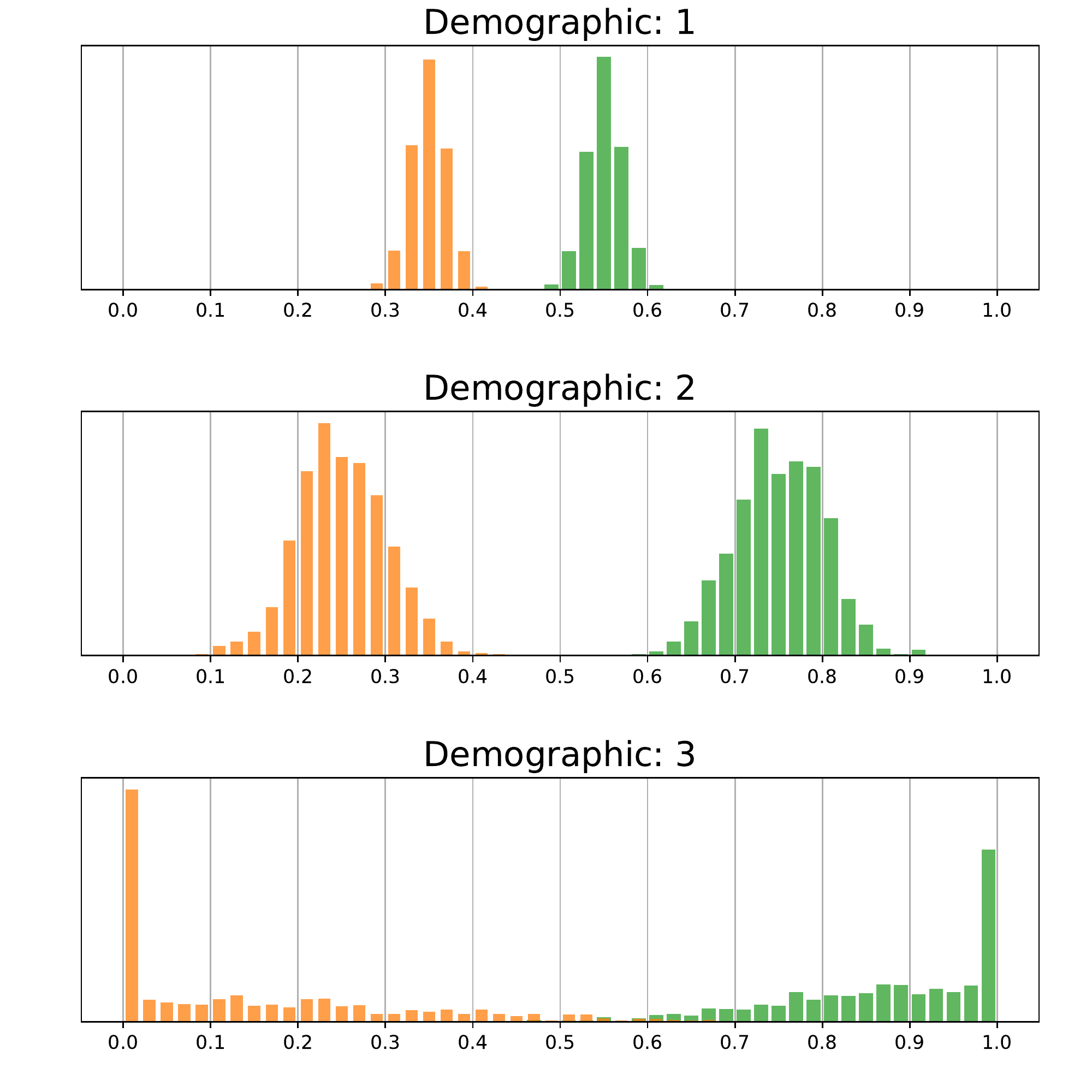}
    \caption{Highly Unfair BV}
    \label{fig:dfi_extreme}
\end{subfigure}
\caption{Score distributions of a canonical biometric verification systems with
        three demographic groups evaluated for DFI measure. Histogram of
        genuine and imposter scores are represented by green and orange bars,
        respectively.}
\label{fig:dfi_n}
\vspace{-4mm}
\end{figure}

The literature is enriched with the similarity measures between probability
distributions. We consider the Jensen-Shannon divergence~\cite{jsd} to quantify
the similarity among score distributions of demographic groups. Unlike previous
two fairness measures, the DFI does not distinguish between the scores of
genuine and imposter pairs for a given demographic group, $d_i$. This measure,
thus, considers the complete data and scores of the demographic group as a
whole, and measures how different these distributions are from their
\textit{mean} distribution. Fig.~\ref{fig:dfi_n} depicts the fair and
unfair verification systems from the perspective of DFI measure. The procedure
to compute the DFI is provided below.
\begin{align}
\label{eq:dfi_n}    
&z_{D_\mathrm{mean}} = \frac{1}{K} \, \sum_{i=1}^{K} z_{D_i} \nonumber \\
&\mathrm{DFI_N} = 1 - \frac{1}{K\, \log_2 K} \sum_{i=1}^{K} D_\mathrm{KL}(z_{D_i} || z_{D_\mathrm{mean}} ),
\end{align}
where $z_{D_i}$ is the combined (genuine + imposter) score distribution,
normalized to yield a unity sum, of the demographic group $d_i$. To compute the
score histogram we empirically consider 100 bins. The Kullback-Leibler (KL)
divergence, $D_\mathrm{KL}$, has been computed on distributions
$A(x)$ and $B(x)$ considering 100-binned score histogram as follows:
\begin{equation}
\label{eq:kld}
D_\mathrm{KL}(A(x) || B(x) ) = \sum_{x} A(x) \log_2{\left(\frac{A(x)}{B(x)}\right)}.
\end{equation}

The normal variant of the DFI measures the summation of average distance
between the score distributions of demographic groups. In the best case
scenario, where the BV system generates nearly same score distributions (as a
whole) for each demographic group, the value of $\mathrm{DFI_N}$ approaches to
one. The extremal variant of DFI computed using the demographic group with
maximum KL divergence from the average distribution. 
\begin{align}
\label{eq:dfi_e}    
&\mathrm{DFI_E} = 1 - \frac{1}{\log_2 K} \, \max_i D_\mathrm{KL}(z_{D_i} || z_{D_\mathrm{mean}} ).
\end{align}

For the weighted variant of DFI, we replace the relative weights ($1/K$) from
normal variant to the sample size-based weights ($w_i$) as discussed in the
previous section. However, we do not modify the computation of average
distribution using fusion weights. The formula for DFI\textsubscript{W} is
provided in Eq.~\ref{eq:dfi_w}.
\begin{align}
\label{eq:dfi_w}    
\mathrm{DFI_W} = 1 - \frac{1}{\log_2 K} \,  \sum_{i=1}^{K} w_i \, D_\mathrm{KL}(z_{D_i} || z_{D_\mathrm{mean}} ).
\end{align}

The values of all variants of the DFI, corresponding to the score distributions
in Fig.~\ref{fig:dfi_n}, are provided below in Table~\ref{tb:dfi_n}. It also
includes the values for demographically imbalanced datasets where only the
weighted variant obtains different values than the balanced scenario.
\begin{table}[h]
\renewcommand{\arraystretch}{1.0}
\centering
\caption{The values of the DFI measure on canonical BV systems. For balanced
        case, 1000 samples were selected for each demographic group. For
        imbalanced case, the number of samples per demographic group was
        modified to [100, 1000, 2000].}
\label{tb:dfi_n}
\begin{tabular}{ l | p{1.5cm} p{1.5cm} p{1.5cm} || p{1.5cm} p{1.5cm} p{1.5cm}}  \hline			
  & \multicolumn{3}{c ||}{Balanced} &\multicolumn{3}{c}{Imbalanced} \\ \cline{2-7}
  & Fair & Unfair & Highly Unfair & Fair & Unfair & Highly Unfair \\ \hline
$\mathrm{DFI_N}$ & 0.9744 & 0.9206 & 0.3367 & 0.9293 & 0.8763 & 0.3364 \\
$\mathrm{DFI_E}$ & 0.9687 & 0.8930 & 0.1295 & 0.8570 & 0.7820 & 0.1198 \\
$\mathrm{DFI_W}$ & 0.9744 & 0.9206 & 0.3367 & 0.9219 & 0.8622 & 0.3052 \\ \hline
\end{tabular}
\end{table}
%
%


\section{Summary}
\label{sec:summary}

In this work, we have introduced three measures for evaluation of demographic
fairness of a generic biometric verification system. The proposed measures
determine the fairness of a verification system, towards demographic groups,
based on its equitability w.r.t separation, compactness, and distribution of
genuine and imposter scores. In addition to mathematical expressions, we have
also provided practical meaning and desired behavior of these fairness
measures. 

We have discussed three variants of each fairness measure based on how the
effect of each demographic group contributes towards the final measure. We have
also addressed the concern related to the demographic imbalance in test
datasets. We have discussed why a simple linear relationship between fusion
weights and sample sizes (of demographic groups) is not effective. Our weighted
fusion strategy attempts to balance relative importance of under-represented
demographic groups without aggravating their contributions towards the final
fairness measure. Such fusion strategies could be useful in comparing the
fairness of biometric systems on imbalanced datasets. 

Our work seeks at decoupling the notion of fairness of a verification system
from its accuracy. Since we employ a differential performance-based approach,
our measures are dependent on the chosen score function, but not on any
external parameters such as score threshold. The proposed measures are not be
considered as alternative to the outcome-based fairness measures, rather both
evaluation approaches are complementary towards analysis of the demographic
fairness of a biometric verification system.
%


\section*{Acknowledgements}
Authors would like to thank the Hasler foundation
for their support through the SAFER project.


\bibliographystyle{ieeetr}
\bibliography{ref1}
\end{document}